\documentclass[11pt,letterpaper]{article}
\usepackage{ijcnlp2017}
\usepackage{times}
\usepackage{latexsym}
\usepackage{graphicx}
\usepackage{mathtools}
\usepackage{amsmath}
\usepackage{enumitem}
\usepackage{wrapfig}
\usepackage{caption}
\usepackage{subcaption}
\ijcnlpfinalcopy

\title{Deep Automated Multi-task Learning}

 \author{Davis Liang \\ d1liang@ucsd.edu \\ University of California, San Diego \AND Yan Shu \\ yashu@ucsd.edu \\ University of California, San Diego}

\date{}

\begin{document}

\maketitle

\begin{abstract}
Multi-task learning (MTL) has recently contributed to learning better representations in service of various NLP tasks. MTL aims at improving the performance of a primary task, by jointly training on a secondary task. This paper introduces automated tasks, which exploit the sequential nature of the input data, as secondary tasks in an MTL model. We explore next word prediction, next character prediction, and missing word completion as potential automated tasks. Our results show that training on a primary task in parallel with a secondary automated task improves both the convergence speed and accuracy for the primary task. We suggest two methods for augmenting an existing network with automated tasks and establish better performance in topic prediction, sentiment analysis, and hashtag recommendation. Finally, we show that the MTL models can perform well on datasets that are small and colloquial by nature.
\end{abstract}

\section{Introduction}

Recurrent neural networks have demonstrated formidable performance in NLP tasks ranging from speech recognition \cite{hinton2012deep} to neural machine translation \cite{bahdanau2014neural,wu2016google}. In NLP, multi-task learning has been found to be beneficial for \textsl{seq2seq} learning \cite{luong2015multi, cheng2016semi}, text recommendation \cite{bansal2016ask}, and categorization \cite{liu2015representation}. 

Despite the popularity of multi-task learning, there has been little work done in generalizing the application of MTL to all sequential tasks. To accomplish this goal, we use the concept of automated tasks. Similar work in multi-task learning frameworks proposed in \cite{liu2016deep} and \cite{luong2015multi} are both trained on multiple labeled datasets. Though we have seen evidence of research using external unlabeled datasets in pre-training \cite{dai2015semi} and semi-supervised multi-task frameworks \cite{ando2005}, to our knowledge there is no work dedicated to using tasks derived from the original dataset in multi-task learning with deep recurrent networks. With automated tasks, we are able to use MTL for almost any sequential task.

\begin{figure}[t!]
    \centering
    \begin{subfigure}[b]{0.4\textwidth}
        \includegraphics[width=\textwidth]{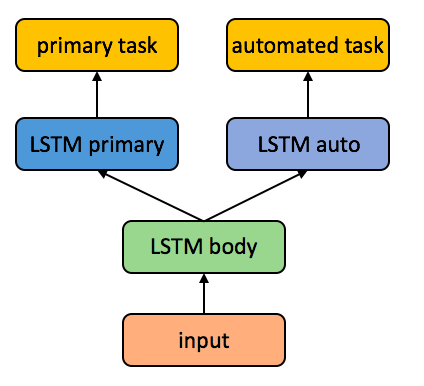}
        \caption{}
        \label{fig:caption-local}
    \end{subfigure}
    \begin{subfigure}[b]{0.4\textwidth}
        \includegraphics[width=\textwidth]{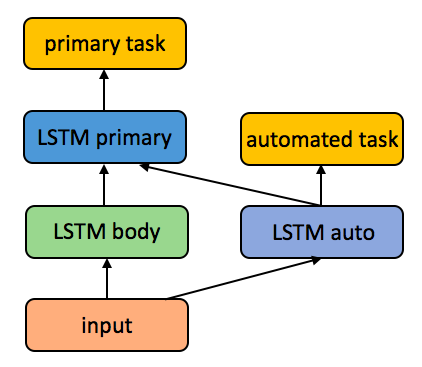}
        \caption{}
        \label{fig:caption-global}
    \end{subfigure}
    \caption{MRNN (a) and CRNN (b) model}\label{fig:both}
\end{figure}

We present two ways of using automated multi-task learning: (1) the MRNN, a multi-tasking RNN where the tasks share an LSTM layer, and (2) the CRNN, a cascaded RNN where the network is augmented with a concatenative layer supervised by the automated task. Examples of either network are shown in Figure \ref{fig:both}.  

In summary, our main contributions are: 
\begin{itemize}[noitemsep]
\item We introduce the concept of automated tasks for multi-task learning with deep recurrent networks. \item We show that using the CRNN and the MRNN trained in parallel on a secondary automated task allows the network to achieve better results in sentiment analysis, topic prediction, and hashtag recommendation.
\end{itemize}

\section{Automated Multi-task Learning}

We generalize multi-task learning by incorporating automated tasks with our two MTL models: the CRNN and the MRNN. In the following subsections, we describe the automated tasks, the models, and their respective training methods. 

\subsection{Automated Tasks}
The set of automated tasks we suggest include (1) next word prediction, (2) next character prediction, and (3) missing word completion.

For word and character generation, we trained a language model to predict the next word or character given the words or characters from the previous K steps. For the missing word completion task, we removed a random non-stop-word from each document and replaced it with a UNK placeholder. The removed word is fed into a word2vec model trained on Google News \cite{le2014distributed} and the resulting vector is the target. We performed regression to minimize the mean squared error of predicting the missing word vector given the text. We generated predictions by finding the target word vector with the highest cosine similarity to the output vector. 


\subsection{MRNN}
The multi-tasking RNN, MRNN, is an MTL model that we use to train our primary and automated tasks in parallel. The MRNN's initial layers are shared, and the later layers branch out to separate tasks. A basic example of an MRNN is shown in Figure \ref{fig:both}. 

The MRNN is constructed such that the primary task and automated task(s) share a body of units. This body is supervised by both the primary and automated task(s) and learns internal representations for both tasks. 

\subsection{CRNN}
\cite{sogaard2016deep} showed that a higher-level task can benefit from making use of a shared representation learned by training on a lower-level task. Similarly, the CRNN assumes that the primary task has a hierarchical relationship with the automated task. A basic example of a CRNN is shown in Figure \ref{fig:both}.

Specifically, we designed the CRNN to use the representations learned from an automated task as a concatenative input \cite{ghosh2016contextual, lipton2015generative} for the primary task. Furthermore, such a model can be supervised on an identical task at different network layers.


\section{Experiments}
\label{sec:experiments}
We evaluate the performance of our models on binary sentiment analysis of the Rotten Tomato Movie Review dataset, topic prediction on the AG News dataset, and hashtag recommendation on a Twitter dataset. For each of these datasets, we compared the results from the MRNN and CRNN to a corresponding LSTM model. We separately tuned the hyper-parameters for each model with the validation sets and took the average results across the multiple runs.  Note that the baseline LSTM models are 2-layered. Our MTL models and the LSTM baseline have the exact same number of parameters along the primary task stream. 

In the following experiments, we use 512 LSTM cells for all models trained on the Rotten Tomato dataset and 128 LSTM cells for the AG News and Twitter datasets. Before each output layer, we have a single fully connected layer consisting of 512 hidden units for the Rotten Tomato dataset and 128 hidden units for the AG News and Twitter datasets. We use a batch size of 128 and apply gradient clipping with the norm set to 1.0 on all the parameters for all experiments. 
 
We found that missing word completion is especially detrimental to our MTL models. We believe that removing a word from each document, which consists almost exclusively of short sequences, discards a large portion of the useful information. Thus, the quantitative results of the missing word completion experiments have been omitted from this paper. We hypothesize that missing word completion is more useful for datasets with longer documents where discarding individual words will not have a major effect on each document.

\subsection{Data}
\label{sec:data} 
The Rotten Tomato Movie Review (RTMR)\footnote{\scriptsize {\tt cs.cornell.edu/people/pabo/movie-review-data/}} \cite{pang2005seeing} dataset consists of 5331 positive and 5331 negative review snippets. The task is to predict review sentiment. The dataset is randomly split into 90\% for the training and validation sets and 10\% for test set ~\cite{dai2015semi}.

The AG News\footnote{\scriptsize {\tt di.unipi.it/gulli/AG\_corpus\_of\_news\_articles.html}} \cite{zhang2015character}
 dataset consists of 120,000 training and 7,600 testing documents. The task is to classify the documents into one of four topics. Following \cite{wangrecurrent}, we took 18,275 documents from the training set as validation data.

The Twitter dataset consists of 5,964 tweets. The task is to predict one of the 71 hashtag labels. We collected 300,000 tweets using the Twitter API. We removed all retweets, URLs, uncommon symbols, and emojis. We lowercased all the characters in the tweets. We kept the tweets with the 71 most popular English hashtags, and removed the hashtags from the tweets. We used an 80/10/10 split of the remaining data. Although Twitter's Developer Policy prevents us from releasing the dataset, the entire data collection pipeline will be made available upon publication.

\begin{table}
\centering
\small
\begin{tabular}{|l|l|l|l|}
\hline
{\bf Dataset} & {\bf Doc. Count} & {\bf Categories} &{\bf Avg. WC} \\\hline
\verb|RTMR| & {10662} & {2} & {20}\\
\verb|AGNews| & {127600} & {4}& {34} \\
\verb|Twitter| & {5964} & {71} & {70*}\\\hline
\end{tabular}
\caption{Dataset statistics. (*character count)}\label{tab:accents}
\vspace{-0.25in}
\end{table}

\section{Rotten Tomatoes}
\subsection{Training Details}
\label{sec:training}
The primary task for the Rotten Tomatoes dataset is sentiment analysis. We used word generation as the automated task. The input is a 300-dimensional word2vec vector for each word. The primary task output consists of two softmax units, representing a positive or negative review. The automated task output is next word prediction of the word2vec representation, and hence is a 300 unit tanh layer. For LSTM we use a learning rate of 0.0001. For the MTL models, we need to tune the learning rate hyper-parameter of the automated task. Instead of tuning the primary and automated task hyper-parameters separately, we found an alternative method for tuning the learning rates using the following equation where $lr_{\text{actual}}$ is the only learning rate hyper-parameter.  $lr_{\text{actual}}$ is optimized on the validation set.

\vspace{-0.2in}
\begin{equation}
\begin{split}
lr_{\text{prim}}(\text{epoch}) = \text{epoch}*(\frac{lr_{\text{actual}}}{\text{total Epochs}})\\
lr_{\text{auto}}(\text{epoch}) = lr_{\text{actual}} - lr_{\text{prim}}(\text{epoch})
\end{split}
\end{equation}
We apply this type of learning rate modulation in order to simulate network pre-training on the automated task in the earlier epochs, learn shared representations in the intermediate epochs through multi-task learning, and train more exclusively on the primary task during the later epochs. We used an $lr_{\text{actual}}$ of 0.01.

The MTL and LSTM models both use word-level word2vec representations trained on Google News \cite{le2014distributed}. The primary sentiment analysis task is trained using Adam optimizer \cite{kingma2014adam} on cross-entropy loss while the automated word generation task is trained using mean-squared error. We continue to use Adam optimizer in the rest of our experiments.

\subsection{Results}
\label{sec:results}
We compare our experimental results with (1) SA-LSTM ~\cite{dai2015semi}, an LSTM initialized with a sequence auto-encoder, and (2) the adversarial model \cite{miyato2016adversarial},  an LSTM-based text classification model with perturbed embeddings. We choose these two models because they are both LSTM-based and are thus comparable to our models. Non-LSTM models, such as convolutional neural networks, have been able to achieve higher accuracy on sentiment analysis with the Rotten Tomatoes dataset \cite{kim2014convolutional}. All of our networks beat the variant of the SA-LSTM that does not use outside data for pre-training. However, the adversarial \cite{miyato2016adversarial} and SA-LSTM \cite{dai2015semi} models, using external unlabeled datasets, outperform our MTL models. With the MRNN, we achieve a 1.5\% gain in accuracy over SA-LSTM, and 1\% over the vanilla LSTM network. With the CRNN, we achieve similar results compared to the vanilla LSTM network. We hypothesize that the reason the CRNN under-performs the MRNN is due to the lack of a clear hierarchy between sentiment analysis and word generation. We suspect that sentiment analysis is primarily keyword based and cannot fully take advantage of the automated language model task. Additionally, we found that the MTL models can be trained with much higher learning rates than a standard LSTM, allowing for convergence in many fewer epochs. The MRNN model converged within the first 10 epochs, whereas the LSTM model required approximately 30 epochs to converge. 

\section{AG News}
\subsection{Training Details}
\label{sec:training}
For the AG News experiment, the primary task is topic prediction and the automated task is word generation. The input to the model is the 300-dimensional word2vec representations of the words from the documents. The primary task output uses a softmax layer with 4 units. The automated task output is represented by a tanh layer with 300 units. The learning rate for the LSTM is 0.001. For the MRNN, the learning rates undergo the same linear function as in the Rotten Tomatoes experiment where $lr_{\text{actual}}$ is 0.01. 

\subsection{Results}
\label{sec:results}
For AG News dataset, we compare our experiment result with skip-connected LSTM \cite{wangrecurrent}, the previous state-of-the-art model on this dataset. The CRNN outperforms state-of-the-art by 0.14\% and MRNN by 0.26\%. We believe the CRNN beats the MRNN due to a hierarchical relationship between topic prediction and word generation. We suspect that topic prediction, which relies on a holistic understanding of a document, can effectively take advantage of the language model.

\begin{table}
\small
\begin{tabular}{|l|l|l|}
\hline
{\bf Dataset} & {\bf Model} & {\bf Accuracy} \\\hline
\verb|RTMR| & {SA-LSTM \shortcite{dai2015semi}} & {79.7\%}\\
\verb|RTMR| & {SA-LSTM \shortcite{dai2015semi}*} & {83.3\%}\\
\verb|RTMR| & {Adversarial \shortcite{miyato2016adversarial}*} & {83.4\%}\\
\verb|RTMR| & {LSTM} & {80.2\%}\\
\verb|RTMR| & {\bf CRNN} & {\bf 80.1\%}\\
\verb|RTMR| & {\bf MRNN} & {\bf 81.2\%}\\\hline
            
\verb|AGNews| & {SC-LSTM-I \shortcite{wangrecurrent}} &  {92.05\%}\\ 
\verb|AGNews| & {LSTM} & {91.59\%}\\
\verb|AGNews| & {\bf CRNN} & {\bf 92.19\%}\\
\verb|AGNews| & {\bf MRNN} & {\bf91.93\%}\\\hline

\verb|Twitter| & {LSTM} & {57.8\%}\\
\verb|Twitter| & {\bf CRNN} & {\bf61.4\%}\\
\verb|Twitter| & {\bf MRNN} & {\bf 62.0\%}\\\hline
\end{tabular}
\caption{Experimental results. (*trained on external unlabeled dataset)}\label{tab:accents}
\end{table}

\section{Twitter}
\label{sec:hostile}
We ran an experiment showing that our models can perform well in challenging environments with little data. We used a small dataset of 5,964 tweets. We performed regression on the word2vec representation of the hashtag given the tweet. We chose regression over classification of one-hot targets because our chosen hashtags are inherently non-orthogonal and can benefit from semantic representations in vector space. We trained three models: an LSTM model, the MRNN, and the CRNN.

\subsection{Training Details}
\label{sec:training}
For the Twitter experiment, the primary task is hashtag recommendation and the automated task is character prediction. We use character prediction as the automated task due to the large amount of misspellings and colloquialisms in tweets.

The input to the model is the 66-dimensional one-hot encoding of the characters corresponding to the ASCII characters that we kept during preprocessing. The primary task output is a tanh layer with 300 units. The automated task output uses a softmax layer with 66 units. For all the models we chose a fixed learning rate of 0.001 based on our observation that different learning rates have little effect on the relative trend between the models on this particular task. A constant, equal learning rate allows us to compare the accuracy curves of each network against epochs run.

Since several of the hashtags are very similar to each other (i.e. \#Capricorn and \#Scorpio), we marked a prediction as correct if the predicted semantic vector's top 5\% (top 4) closest cosine distance words contained the target hashtag. 

\subsection{Results}
\label{sec:results}
\begin{figure}[t]
  \centering
  \includegraphics[keepaspectratio, width=0.42\textwidth]{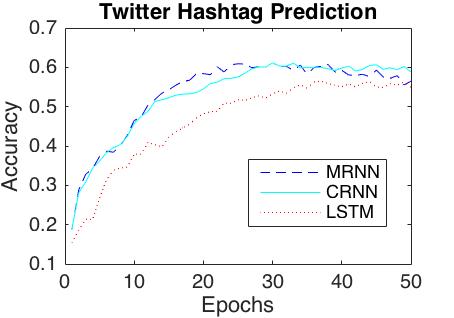}
  \caption{Hashtag prediction in Twitter.}
  \label{fig:tweetnet}
\end{figure}

With the MRNN, we achieve a 4.2\% gain in accuracy over the LSTM in the Twitter dataset. With the CRNN, we achieve a 3.6\% gain in accuracy. Additionally, we have shown in Figure \ref{fig:tweetnet} that both the MRNN and CRNN models converge faster than the LSTM model; both MTL models take approximately half of the number of epochs to reach 50\% accuracy using the same constant learning rate.

\section{Conclusion }
\label{sec:conclusion}
In this paper, we showed that automated multi-task learning models can consistently outperform the LSTM in sentiment analysis, topic prediction, and hashtag recommendation. Note that the concept of automated tasks can be extended to non-NLP sequence tasks such as image categorization with next row prediction as the automated task. Because automated MTL can be integrated into an existing network by adding a new branch to a pre-existing graph, we can substitute bidirectional LSTMs \cite{schuster1997bidirectional}, GRUs \cite{34012d808c814b5593264c1037ed3dc6}, and vanilla RNNs for LSTMs in our MTL models. We will experiment on these variations in the future.

\bibliography{ijcnlp2017}
\bibliographystyle{ijcnlp2017}

\end{document}